\documentclass{article}

\usepackage{microtype}
\usepackage{graphicx}
\usepackage{booktabs} 

\usepackage{hyperref}

\usepackage[accepted]{icml2025}

\usepackage{amsmath}
\usepackage{amssymb}
\usepackage{mathtools}
\usepackage{amsthm}
\usepackage{bbm}

\usepackage{graphicx}
\usepackage{tabularray}
\usepackage{wrapfig}
\usepackage{bbm}
\usepackage{enumitem}
\usepackage{amsmath,amssymb,amsthm}
\usepackage{adjustbox}
\usepackage{array}
\usepackage{multicol}
\usepackage{multirow}
\usepackage{makecell}
\usepackage{booktabs}
\usepackage{pifont}
\usepackage{subfig}
\usepackage{caption}
\usepackage[most]{tcolorbox}
\usepackage{thmtools}
\usepackage{fontawesome5}
\usepackage{algorithm,algorithmic}
\usepackage{listings}
\usepackage{booktabs,siunitx}

\usepackage[capitalize,noabbrev]{cleveref}

\theoremstyle{plain}
\newtheorem{theorem}{Theorem}[section]

\newtheorem{example}[theorem]{Example}
\theoremstyle{definition}
\newtheorem{definition}[theorem]{Definition}

\theoremstyle{remark}

\usepackage[textsize=tiny]{todonotes}

\usepackage{amsmath,amsfonts,bm}

\def\eqref#1{equation~\ref{#1}}

\def\1{\bm{1}}

\DeclareMathAlphabet{\mathsfit}{\encodingdefault}{\sfdefault}{m}{sl}
\SetMathAlphabet{\mathsfit}{bold}{\encodingdefault}{\sfdefault}{bx}{n}

\def\gD{{\mathcal{D}}}
\def\gE{{\mathcal{E}}}

\def\gI{{\mathcal{I}}}

\def\gL{{\mathcal{L}}}

\def\gN{{\mathcal{N}}}

\def\gY{{\mathcal{Y}}}
\def\gZ{{\mathcal{Z}}}

\def\sP{{\mathbb{P}}}

\def\sR{{\mathbb{R}}}

\newcommand{\E}{\mathbb{E}}

\newcommand{\R}{\mathbb{R}}

\DeclareMathOperator*{\argmax}{arg\,max}

\icmltitlerunning{Few-shot Steerable Alignment: Adapting Rewards and LLM Policies with Neural Processes}

\begin{document}

\twocolumn[
\icmltitle{Few-shot Steerable Alignment:\\Adapting Rewards and LLM Policies with Neural Processes}



\icmlsetsymbol{equal}{*}

\begin{icmlauthorlist}
\icmlauthor{Katarzyna Kobalczyk}{equal,yyy}
\icmlauthor{Claudio Fanconi}{equal,yyy}
\icmlauthor{Hao Sun}{yyy}
\icmlauthor{Mihaela van der Schaar}{yyy}

\end{icmlauthorlist}

\icmlaffiliation{yyy}{Department of Applied Mathematics and Theoretical Physics, University of Cambridge, Cambridge, United Kingdom}

\icmlcorrespondingauthor{Katarzyna Kobalczyk}{knk25@cam.ac.uk}
\icmlcorrespondingauthor{Claudio Fanconi}{caf83@cam.ac.uk}

\icmlkeywords{Machine Learning, ICML}

\vskip 0.3in
]



\printAffiliationsAndNotice{\icmlEqualContribution} 

\begin{abstract}
As large language models (LLMs) become increasingly embedded in everyday applications, ensuring their alignment with the diverse preferences of individual users has become a critical challenge. Currently deployed approaches typically assume homogeneous user objectives and rely on single-objective fine-tuning. However, human preferences are inherently heterogeneous, influenced by various unobservable factors, leading to conflicting signals in preference data. Existing solutions addressing this diversity often require costly datasets labelled for specific objectives and involve training multiple reward models or LLM policies, which is computationally expensive and impractical. In this work, we present a novel framework for few-shot steerable alignment, where users' underlying preferences are inferred from a small sample of their choices. To achieve this, we extend the Bradley-Terry-Luce model to handle heterogeneous preferences with unobserved variability factors and propose its practical implementation for reward modelling and LLM fine-tuning. Thanks to our proposed approach of functional parameter-space conditioning, LLMs trained with our framework can be adapted to individual preferences at inference time, generating outputs over a continuum of behavioural modes. We empirically validate the effectiveness of methods, demonstrating their ability to capture and align with diverse human preferences in a data-efficient manner. Our code is made available at: \href{https://github.com/kasia-kobalczyk/few-shot-steerable-alignment}{https://github.com/kasia-kobalczyk/few-shot-steerable-alignment}.
\end{abstract}

\section{Introduction}


\textbf{Motivation: The need for personalisation.} As large language models (LLMs) become more integrated into daily life, it is critical that these models align with diverse human values and preferences. Current methods for LLM training, such as Reinforcement Learning from Human Feedback (RLHF) and Direct Preference optimisation (DPO), rely on single-objective fine-tuning, which assumes uniformity in human preferences. In reality, users of LLM-based systems often have diverse and even conflicting goals, influenced by factors like demographic and cultural backgrounds, cognitive processes, or implicit ambiguity in their prompt messages. These factors are often unobservable, leading to preference data that contains conflicting signals. When such data is aggregated under a homogeneous Bradley-Terry-Luce (BTL) model \citep{bradley_rank_1952}, the result is equivalent to a Borda count \citep{siththaranjan_distributional_2024}, which fails to capture the true diversity in human preferences.

\textbf{Challenges: Unknown sources of heterogeneity require new methods.} To overcome the limitations of single-objective fine-tuning, researchers have explored training multi-objective language models that generate desirable outputs by aligning with weighted mixtures of reward functions \citep{guo_controllable_2024, wang_arithmetic_2024, rame_rewarded_2023, jang_personalized_2023, wang_conditioned_2024}. However, these methods typically require multiple datasets, each labelled for a specific objective (e.g., honesty, helpfulness, or safety), and expect users to represent their preferences as a weighted vector over predefined objectives. In cases where the sources of preference heterogeneity are unobservable, this approach becomes impractical. Additionally, existing approaches require training multiple independent reward models or even various LLM policies \citep{chidambaram_direct_2024}, one for each objective, which can be prohibitively expensive regarding computational resources and storage.

\textbf{Our goal: Towards few-shot steerable alignment.} In this paper, we address the challenge of few-shot steerable alignment: Can we infer a user's underlying preferences from a small \textit{few-shot} sample of their preference choices, and can we train LLMs to align with these preferences at \textit{inference time}? Our goal is to achieve this alignment based on a dataset of heterogeneous preference choices without relying on predefined objectives or requiring training multiple independent reward or language models.

\textbf{Contributions:} $\blacktriangleright$  We extend the standard BTL model to account for the heterogeneous preferences across a population of users, enabling few-shot adaptation to individualised reward functions. $\blacktriangleright$  By modelling user-level reward functions as samples from a stochastic process, we propose a practical implementation of this model based on Neural Processes (NPs) \citep{garnelo_conditional_2018, garnelo_neural_2018}. We demonstrate its applicability in both reward modelling (\textbf{NP-BTL}) and direct preference optimisation (\textbf{NP-DPO}). $\blacktriangleright$ By introducing functional, parameter-space conditioning, NP-DPO opens up the possibility for training LLMs that be adapted to individual users' preferences at inference time and can generate outputs across a \textit{continuum} of behavioural modes, reflecting diverse human preferences without constraining the factors of variability to a finite set of fixed, pre-defined objectives nor requiring training and storage of multiple models. $\blacktriangleright$  We empirically evaluate NP-BTL and NP-DPO, demonstrating and critically analysing their effectiveness in capturing and aligning with heterogeneous human preferences.

\section{Background \& Problem Definition}\label{sec:background_problem_defintion}

\subsection{Background}\label{subsec:background}

\textbf{Standard setup.} Throughout this paper, we let $\gY$ denote the space of candidate options that users rank according to their preferences. For any pair of options $y_1, y_2 \in \gY$ we use the notation $y_1 \succ y_2$ to denote that $y_1$ is preferred over $y_2$. Conventional approaches to preference learning and optimisation start by collecting a dataset of user's preference choices $\gD$, represented as tuples, $\gD = \{(y^w_j, y^\ell_j)\}$ with $y^w_j, y^\ell_j \in \gY$ such that $y_j^w \succ y^\ell_j$. In the context of natural language modelling, $\gY$ is the space of natural language. Users express their preferences over pairs $y_1, y_2 \in \gY$, with $y_1 = [x, \tilde{y}_1]$,  $y_2 = [x, \tilde{y}_2]$, where $x$ is a prompt common for both options and $\tilde{y}_1, \tilde{y}_2$ two distinct completions sampled from an LLM, denoted by $\tilde{y}_1, \tilde{y}_2 \sim \pi_\theta(\cdot \vert x)$, where $\pi_\theta(\cdot \vert x)$ is the generative distribution of the LLM, aka the LLM policy.

A standard RLHF procedure starts by learning a reward function $r: \gY \rightarrow \mathbb{R}$ aiming to capture the utility of individual options. The higher the utility of a candidate option $y$, the more likely it is that it is preferred over the alternatives. After fitting the reward function to the dataset $\gD$, it is then used to optimise the LLM policy $\pi_\theta$ via supervised fine-tuning. Alternatively, these two steps can be combined into one optimisation step, leading to the Direct Preference Optimisation (DPO) \citep{rafailov_direct_2024} procedure. We outline the details below.

\textbf{RLHF under the BTL model.} The Bradley-Terry-Luce model \cite{bradley_rank_1952} (BTL) is the most common approach to modelling user preferences. It models the likelihood of $y_1 \succ y_2$ as:
\begin{align*}
    p(y_1 \succ y_2) &=  \sigma(r(y_1) - r(y_2)) \\&:= \frac{\exp(r(y_1))}{\exp(r(y_1)) + \exp(r(y_2))},
\end{align*}
where $r: \gY \rightarrow \sR$ is a reward function assumed common for all users. Typically, $r$ is modelled by a neural network $r_\phi$ whose parameters $\phi$ are fitted by maximising the likelihood procedure on the preference dataset $\gD$. The learned reward function $r_\phi$ is then used to train the LLM policy $\pi_\theta$ by maximising:
\begin{equation*}
    \max_{\theta} \E_{x \sim \gD, \tilde{y} \sim \pi_\theta(\cdot \vert x)}\left[r_\phi(y)\right] - \beta\mathrm{D}_{KL}[\pi_\theta(\tilde{y} \vert x) || \pi_{\mathrm{ref}}(\tilde{y} \vert x)],
\end{equation*}
where $\beta$ is a parameter controlling the deviation from the base reference policy $\pi_{\mathrm{ref}}$. 

\textbf{DPO under the BTL model.} By using a change of variables, DPO directly optimises the parameters $\theta$ according to the following objective:
\begin{equation*}
\scalebox{0.85}{$  
    \displaystyle
    \max_{\theta} \mathop{\E}_{(x, \tilde{y}^w, \tilde{y}^\ell) \sim \gD}\left[\log \sigma\left(\beta \log \frac{\pi_\theta(\tilde{y}^w \vert x)}{\pi_{\mathrm{ref}}(\tilde{y}^w \vert x)} - \beta \log \frac{\pi_\theta(\tilde{y}^\ell \vert x)}{\pi_{\mathrm{ref}}(\tilde{y}^\ell \vert x)}\right)\right].
    $}
\end{equation*}

\textbf{Distributional Preference Learning.} We note that both methods of RLHF and DPO based on the BTL model assume that the internal preferences of individual users can be summarised well with a single reward function $r$. As discussed in detail by \cite{siththaranjan_distributional_2024}, this model does not cater to diverse, potentially conflicting user preferences that may stem from unobservable effects of the so-called ``hidden context''. Rewards fitted under the BTL model implicitly aggregate over hidden contexts according to a Borda count rule, which may lead to counter-intuitive results. To uncover the impact of the hidden context, \cite{siththaranjan_distributional_2024} propose distributional preference learning (DPL)--a class of models that estimate a distribution of possible reward values for each option rather than returning simple point estimates. In particular, their proposed mean-variance model assumes that:
\begin{gather*}
    p(y_1 \succ y_2) = \E\left[\mathbbm{1}\{r(y_1) - r(y_2) > 0\}\right],
\end{gather*}
where $r(y) \sim \gN(\mu(y), \sigma(y))$, with both $\mu$ and $\sigma$ modelled as outputs of a neural network. While this model enables the measurement of the overall variability of utilities that users assign to a single option $y$, this model does not solve the problem of identifying the expected value of the utility assigned to a given option for a given individual.

\subsection{Problem Definition}\label{subsec:problem_definition}

\textbf{Our Setup.} With the focus on modelling preferences of individual users, our setup assumes access to a pre-collected dataset $\gD = \{\gD_i\}_{i\in \gI}$ of preference choices collected from multiple users indexed by $i \in \gI$. Each user-specific dataset, $\gD_i$ consists of a number of preference pairs represented as tuples $\gD_i = \{(y_{i,j}^w, y_{i,j}^\ell)\}_{j=1}^{N_i}$, where $y_{i,j}^w$ is the option chosen by user $i$ and $y_{i,j}^\ell$ the rejected option.

\textbf{The goal.} Given a new user $i$, and a small, few-shot sample of their preferences expressed in a context dataset $\gD^C_i = \{(y_{i,j}^w, y_{i,j}^\ell)\}_{j=1}^{N^C}$, we wish to: 
\begin{enumerate}
    \item For any two new options $y_1, y_2 \in \gY$ that are not part of $\gD^C_i$, model the probability of the user preferring $y_1$ over $y_2$.
    \item In the context of language modelling, align the LLM policy $\pi_\theta$ with the user-specific preferences so that for any new prompt $x$, with a high likelihood, the LLM generates content that aligns with the user's choice. 
\end{enumerate}

\section{Method}
The following section outlines our approach to addressing the problem of few-shot steerable alignment defined above. First, in section~\ref{stochastic-reward-modelling}, we extend the BTL reward model to capture user-specific preferences by conditioning on latent variables inferred from the few-shot sample of user-specific contextual data points. We then propose a practical implementation of this model based on Neural Processes \citep{garnelo_conditional_2018, garnelo_neural_2018}--a class of meta-learning models enabling to infer the posterior predictive distribution over the outcomes in a fully amortised manner. We refer to this conditional reward model as NP-BTL. Given our conditional user preference model, we apply it in the context of LLM policy fine-tuning. In section~\ref{conditional-dpo}, we introduce NP-DPO. By extending the conditional setup to DPO, we bypass explicit reward model training and propose a method for directly optimising LLMs with respect to the heterogeneous mixture of user preferences data.  

\subsection{Stochastic Reward Modelling}\label{stochastic-reward-modelling}

We model the heterogeneity of human preferences resulting from the effect of the unobserved hidden context by assuming the following model. We assume that users, indexed by $i$, express their preferences according to their internal reward function $r_i: \gY \rightarrow \mathbb{R}$. We model these functions as samples from a stochastic process. That is, each $r_i = r(\cdot; z_i^*)$ for a particular value of $z_i^* \in \gZ$, where $\gZ$ is a latent sample space. Given a particular value of $z_i^*$ and two options $y_1, y_2 \in \gY$, we model the distribution of preferences extending the BTL model as:
\begin{align}
    p(y_1 \succ y_2 \vert z_i^*) &= \frac{\exp(r(y_1; z_i^*))}{\exp(r(y_1; z_i^*)) + \exp(r(y_2; z_i^*))} \\ &= \sigma\left(r(y_1; z_i^*) - r(y_2; z_i^*)\right).
\end{align}

To address the problem of predicting user preferences for any two previously unobserved options $y_1, y_2$, given a contextual dataset $\gD^C_i$, we will take inspiration from the line of work on Neural Processes (NPs) \cite{garnelo_conditional_2018, garnelo_neural_2018}. An NP is a meta-learning model that learns to map contextual observations to the posterior predictive. In our context, we will aim to approximate:
\begin{equation}
    p(y_1 \succ y_2 \vert \gD^C_i) = \int  p(y_1 \succ y_2 \vert z_i^*)p(z_i^* \vert \gD^C_i)dz_i^*,
\end{equation}
where  $p(z_i^* \vert \gD^C_i)$ is the true conditional of $z_i^*$ given the observed context dataset $\gD^C_i$. NPs approximate the unknown $p(\cdot \vert \gD^C_i)$  with either a variational distribution $q(\cdot \vert \gD^C_i)$ or a deterministic map from the context dataset $\gD^C_i$ to a fixed embedding $z \in \R^d$. For simplicity, we choose to present here the latter option and we model the value of the individual reward function, conditioned on the contextual dataset $\gD^C_i$ as:
\begin{align}\label{eq:conditional-rewards}
    r_{\phi}(y \vert \gD^C_i) &= g_{\phi_d}(y, z_i^C), \\ z_i^C =  h_{\phi_e} (\gD^C_i) &= h_{\phi_{e, 2}}\left(\frac{1}{N^C}\sum_{j=1}^{N^C}h_{\phi_{e, 1}}(y_{i,j}^w, y_{i,j}^\ell)\right),
\end{align}
where $g_{\phi_d}$ and $h_{\phi_e}$ are decoder and encoder networks, respectively. As is standard in NPs, the network $h_{\phi_e}$ is implemented as a Deep Set \cite{zaheer_deep_2018}. This enables us to condition the reward function on an arbitrary number of observed data points $N^C$. We thus have:
\begin{align}
    p(y_1 \succ y_2 \vert \gD^C_i)  &\approx  p_\phi(y_1 \succ y_2 \vert \gD^C_i) \\& = \sigma(r_{\phi}(y_1 \vert \gD^C_i) - r_{\phi}(y_2 \vert \gD^C_i))
\end{align}
We refer to this model as the NP-BTL reward model.

\begin{minipage}{0.45\textwidth}
\begin{algorithm}[H]
\caption{NP-BTL reward model training}
\label{alg:reward-training}
\small
\begin{algorithmic}
    \REQUIRE The dataset $\gD = \{\gD_i\}_{i \in \gI}$, Max iter $T$, Initial parameters $\phi$, Learning rate $\eta$
    \STATE $t = 0$
    \WHILE{ $t < T$}
    \STATE Randomly sample a user $i$
    \STATE Subsample $\gD^C_i$, $\gD^T_i$ from $\gD_i$
    \STATE $\gL(\phi) \leftarrow -\sum_{(y^w, y^l) \in \gD_i^T}\log p_\phi(y^w \succ y^l \vert \gD_i^C)$
    \STATE Update parameters with SGD: $\phi \leftarrow \phi -  \eta \nabla \gL(\phi)$
    \STATE $t \leftarrow t + 1$
    \ENDWHILE
    \STATE \textbf{return} $\phi$
\end{algorithmic}
\end{algorithm}   
\end{minipage}

\begin{figure*}[t]
    \centering
    \subfloat[0.5\textwidth][
    Unconditional reward models: BTL and DPL fitted to the synthetic data vs. the conditional NP-BTL model, given ten randomly sampled contextual data points with $z^*=0$ and $z^*=1$.
    \label{fig:synthetic}
    ]{
    \includegraphics[width=0.6\textwidth]{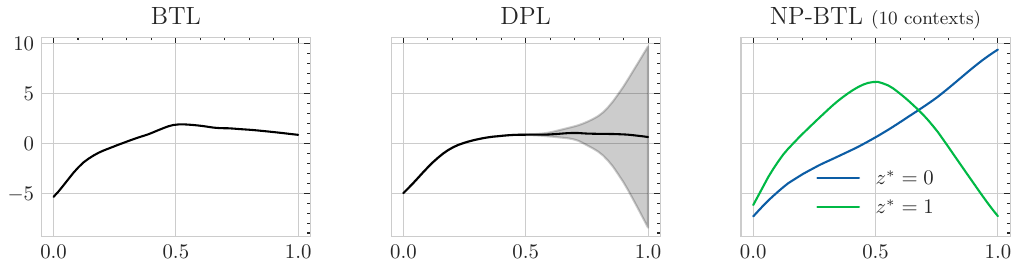}
    }
    \qquad
    \subfloat[0.5\textwidth][
    Accuracy of the NP-BTL model vs. the number of contextual data points. Baseline accuracy is 75\%.
    \label{tab:synthetic}
    ]{
    \hspace{1em}
    \raisebox{0.5\totalheight}{
    \footnotesize
    \begin{tabular}{ll}
    \toprule
    $N^C$ & Accuracy [\%] \\
    \midrule
    0 & 75.2 ± 0.6 \\
    1 & 80.0 ± 0.4 \\
    3 & 88.2 ± 0.3 \\
    5 & 93.6 ± 0.2 \\
    10 & 97.2 ± 0.2 \\
    \bottomrule
    \end{tabular}
    }
    \hspace{1em}
    }
    \caption{\textit{Results on the synthetic data}. Conditioning on user-provided preferences enables the conditional reward model to recover the correct mode of behaviour. As the number of context data points increases}
\end{figure*}

\textbf{Training.} To train the conditional reward model, we find the parameters $\phi$ by maximising the likelihood on batches consisting of pairs of so-called context and target datasets $(\gD^C_i, \gD^T_i)$, where $\gD^C_i$ and $\gD^T_i$ are subsets of preference data obtained from a single user. $\gD^C_i$ is the observed subset of context data points, and $\gD^T_i = \{(y^w_{i,j}, y^\ell_{i,j})\}_{j=1}^{N_T}$ contains pairs of options whose preference order we aim to predict. The simplified training procedure (assuming a batch size of 1) is presented in Algorithm~\ref{alg:reward-training}. An analysis of the complexity can be found in Appendix~\ref{app:algorithm_analysis}.

\subsection{Conditional Direct Preference Optimisation}\label{conditional-dpo}

Our next step is to consider the training of personalisable LLM policies $\pi_\theta$ that can be conditioned on the contextual datasets $\gD^C_i$. To do this, we need to find a way of modulating the output of a language model on a set of preference pairs obtained in $\gD^C_i$, and crucially, be able to do so at inference time. We consider a prototype method based on modulating hypernetworks inspired by \cite{perez_film_2017} to enable this. Namely, we model conditional policies with:
\begin{equation*}
    \pi_\theta(\tilde{y} \vert x, \gD^C_i) = \pi_{\theta'}(\tilde{y} \vert x), \quad \theta' = \theta \odot \tau_i, \quad \tau_i = h_\phi(\gD^C_i).
\end{equation*}
In the above, $\theta$ are the original parameters of the LLM, $\tau_i$ are modulating parameters which are output by the modulating hypernetwork $h_\phi$, again implemented as a Deep Set analogously to (\ref{eq:conditional-rewards}). Details on implementing the modulating operation $\theta \odot \tau_i$ are presented in the Appendix~\ref{app:dpo_implementation}. Extending the conventional DPO objective to our conditional formulation, we optimise the parameters $\theta$ and $\phi$ with respect to:
\begin{equation}\label{eq:dpo-loss}
    \max_{\theta, \phi}\mathop{\mathbb{E}}_{\substack{(\gD^C_i, \gD^T_i),\\ (x, \tilde{y}^w, \tilde{y}^\ell) \in \gD^T_i}}\left[\log \sigma\left(r_\theta(y^w) -  r_\theta(y^\ell)\right)\right],
\end{equation}
where $r_{\theta}(y) := \beta \log \frac{\pi_\theta(\tilde{y} \vert x, \gD^C_i)}{\pi_{\mathrm{ref}}(\tilde{y} \vert x)}$ is the so-called implicit DPO reward~\cite{rafailov_direct_2024}. During training the subsets $(\gD^C_i, \gD^T_i)$ are sampled analogously to Algorithm~\ref{alg:reward-training}. The model learns to condition its outputs on the observed contexts by training over multiple pairs of context and target datasets. The introduced parameter-space conditioning based on FiLM layers enables flexible modulation of the behaviour of the LLM policy. We refer to this training pipeline as NP-DPO.

\begin{figure*}[t]
    \centering
    \subfloat[0.5\linewidth][
        Accuracies of the baseline BTL models (left) and the NP-BTL model (right).
        \label{fig:hh-reward}
    ]{
        \includegraphics[width=0.47\linewidth]{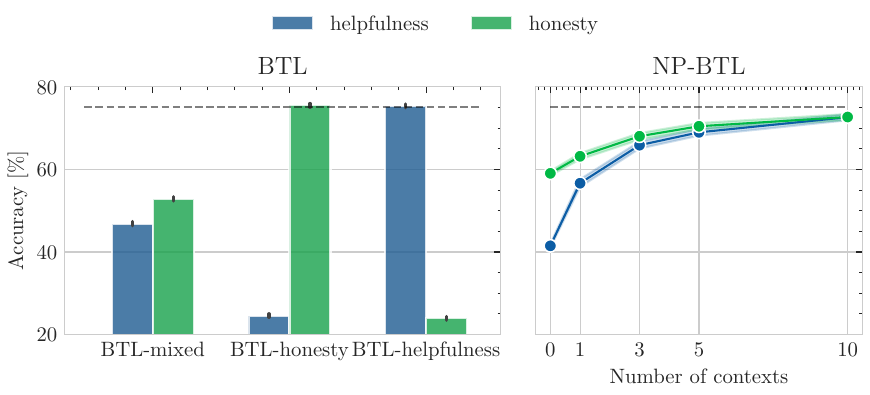}
    }
    \quad
    \subfloat[0.5\linewidth][
        Accuracies of the implicit rewards of LLM policies trained with BTL-DPO (left) and NP-DPO (right).
        \label{fig:hh-policy}
    ]{
        \includegraphics[width=0.47\linewidth]{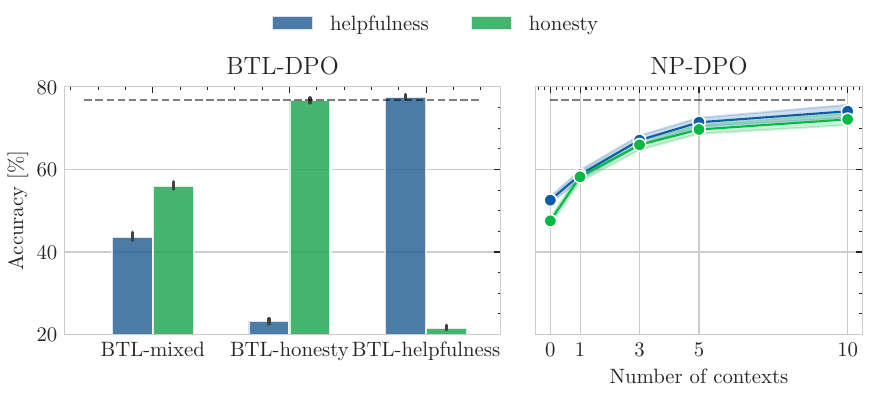}
    }
    \caption{\textit{Performance of a) NP-BTL and b) NP-DPO vs. their non-conditional counterparts.}}
    \label{fig:hh-results}
    \vspace{-1em}
\end{figure*}

\section{Empirical Studies and Insights}\label{sec:experiments}

In this section, we demonstrate the performance of both NP-BTL and NP-DPO on a series of illustrative experiments with synthetic data and larger-scale experiments on an NLP dataset for LLM alignment. We also provide further insights into the efficacy of conditioning with respect to the conflict rate within the training dataset and the contextual examples. The error bars in all plots represent the standard errors across testing tasks. Details on the experimental setups, including the number of training and testing tasks, can be found in Appendix~\ref{sec:appdx-hh-details}. 

\subsection{Illustrative comparison of existing approaches.}

We begin with an illustrative example qualitatively comparing the NP-BTL model with the standard BTL and DPL models. The below presented follows closely that introduced in \cite{siththaranjan_distributional_2024}.
\begin{example}\label{example:synthetic}
    We simulate a heterogeneous population of users expressing their preferences with respect to options represented as values $y \in [0, 1]$. We assume that our population consists of two kinds of users represented by the non-observable latent variables $z^* \sim \mathrm{Bernoulli}(0.5)$ and we model the corresponding reward functions as:
\begin{align}
    r(y, z^*) &= x  \ \text{ for } \ y \leq 0.5 
    \quad  \text{and} \\ 
    r(y, z^*) &= \begin{cases}
        2y & \text{if } z^* = 0 \\
        1 - y & \text{if } z^* = 1
    \end{cases} 
    \ \text{ for } y > 0.5
\end{align}
\end{example}
We generate a synthetic dataset of options $y_1, y_2 \in [0, 1]$ and user choices made according to $p(y_1 \succ y_2) = \mathbbm{1}\{(y_1, z^*) > r(y_2, z^*)\}$ and fit all three models: NP-BTL, BTL, and DPL. During training, the number of contexts for the NP-DPL models varies between 0 and 10, and the number of targets is set to 20. Figure~\ref{fig:synthetic} illustrates the fitted reward functions for all three models. We observe that, as expected, the BTL model averages out the rewards across the entire population, resulting in a flat reward function for $y > 0.5$ and failing to align with the preferences of the individual subgroups of users. While the DPL model can appropriately capture the variation in user preferences for options $y > 0.5$, it cannot adapt to user-specific preferences. Our conditional model successfully recovers the shape of the ground truth rewards. Table~\ref{tab:synthetic} shows the efficacy of the adaptation based on the number of preference choices provided as contexts. We observe that for $N^C = 0$, the conditional model matches the baseline performance of the non-conditional models, $\approx$75\%. As the number of contextual data points increases, so does the alignment of the predicted rewards and the resulting predictions, reaching near-perfect accuracy at $N^C = 10$.


\subsection{Real-world example: Helpfulness vs. Honesty}\label{sec:exp-hh}

We now depart from the controlled setup of synthetic data to demonstrate the applicability of our NP-BTL and NP-DPO models on a real-world UltraFeedback dataset\footnote{\href{https://huggingface.co/datasets/openbmb/UltraFeedback}{https://huggingface.co/datasets/openbmb/UltraFeedback}}\cite{cui_ultrafeedback_2024} commonly used to evaluate RLHF and DPO pipelines. Each sample within this dataset consists of user prompts $x$, two candidate LLM responses $\tilde{y}_1$ and $\tilde{y}_2$ and scores assessing both options with respect to the following categories: helpfulness, honesty, truthfulness and instruction following. To simulate users with varying preferences, we define two reward functions: $r(y; 0) = \textrm{helpfulness}(\tilde{y})$ and $r(y; 1) = \textrm{honesty}(\tilde{y})$. We sample tasks $(\gD^C_i, \gD^T_i)$ during training and testing. At training time, the number of context preference pairs for reward modelling varies between 0 and 10, with the number of targets set to 20. The number of contexts for policy fine-tuning with DPO varies between 0 and 6, with the number of targets set to 8--the reduced number of data points for policy training results from increased memory requirements. At test time, we set the number of targets to 20. Choices between two options $y_1$, $y_2$ are made according to $r(\cdot ; z^*)$ for $z^* \sim \textrm{Bernoulli}(0.5)$. Note, during training and testing, the reward models are only presented with the two options in a text format and the user choice. The underlying scores of helpfulness and honesty are part of the hidden context. In short, we refer to the generated dataset as the HH dataset (not to be confused with the Helpful-Harmless dataset~\cite{Bai2022TrainingAH}).

This experiment aims to investigate the ability of the NP-based reward and policy models to identify the two distinct modes of user preferences and quickly adapt to one of them, given a small sample of contextual preference choices. Thus, we filter our training and testing preference dataset to consist of only pairs which are \textit{conflicting} with respect to $r(\cdot ; 0)$ and $r(\cdot ; 1)$. 

\begin{definition}[Conflicting pairs]
    Given two reward functions: $r_0$, $r_1$, we say that two options $y_1, y_2$ are conflicting if the decision made according to $r_0$ differs from the decision made with $r_1$, i.e. $\mathbbm{1}\{r_0(y_1) > r_0(y_2)\} \neq \mathbbm{1}\{r_1(y_1) > r_1(y_2)\}$.
\end{definition}

Such pairs maximise the information gain with respect to the true value of $z^*$. Section \ref{sec:conflicting} provides further theoretical and empirical insights on the interplay between correlations in the underlying data, the resulting rate of conflicting pairs, and the ability of the model to identify the underlying reward function.

We fit the conditional reward model, NP-BTL, and we also fine-tune the gemma-2b\footnote{\href{https://huggingface.co/google/gemma-2b}{https://huggingface.co/google/gemma-2b}} language model with NP-DPO on our generated dataset. Implementation details can be found in Appendix~\ref{app:dpo_implementation}. 

We compare the accuracies of the NP-BTL reward model and the accuracies of the implicit rewards of NP-DPO with respect to the BTL reward models' accuracies and the implicit rewards of the LM policy trained with standard BTL-DPO. We train two kinds of baseline models: 1) trained on the same heterogeneous dataset of preferences as the NP-based models (BTL-mixed) and 2) trained on each of the two homogeneous subsets of data generated according to the honest and helpful reward functions (BTL-honesty and BTL-helpfulness). Note that the BTL-honesty and BTL-helpfulness models require privileged access to the value of the hidden variables $z^*_i$. Their performance is reported for comparative purposes as the theoretically achievable upper bound.

\begin{figure}[h]
    \centering
    \vspace{-0.5em}
    \includegraphics[width=\linewidth]{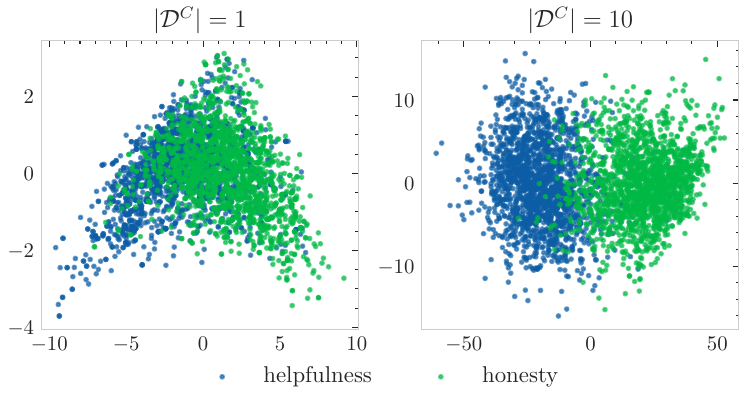}
    \caption{\textit{PCA of context embeddings}. Points are labelled by the true value of $z^*_i$. We observe a clear clustering effect with increasing contextual data points.}
    \label{fig:hh-pca}\vspace{-0.5em}
\end{figure}

Figure~\ref{fig:hh-results} shows the results with models evaluated separately on the honesty valuing and helpfulness valuing subsets of users. We observe the following. Due to the nature of this dataset with 100\% conflicting pairs, the standard BTL-mixed models fail to align with the heterogeneous preferences of the population, resulting in an average performance of ~50\% across the entire population. In contrast, the NP-based rewards and policies successfully adapt to the preferences of individual subgroups as the number of contextual data points increases. At $N^C=10$, the conditional NP-BTL rewards nearly match the accuracies of the dedicated BTL-honesty and BTL-helpfulness models (see dashed lines). Similar observations can be drawn for the NP-DPO implicit rewards. 

We investigate the latent embeddings $z^C_i$ of the context datasets $\gD^C_i$ used to condition the reward functions. In Figure~\ref{fig:hh-pca}, we display the plots of the embeddings reduced to 2 dimensions via principle component analysis (PCA). Points are labelled according to the ground truth value of $z^*_i$ according to which the context dataset has been generated. Two clusters are visible in the graphic, indicating that the model learned to separate the two types of users.

The Appendix section~\ref{app:additional_experiments} contains additional experiments with more than two behavioural modes (Appendix~\ref{app:HHT}) and on a continuum of behaviours (Appendix~\ref{app:continuous_IDs}).

\subsection{Correlations in data and identifiability of the hidden context.}\label{sec:conflicting}

\subsubsection{Analytic study}

In the previous experiment, the training and testing datasets were constructed such that the competing options were perfectly conflicting, i.e. if a user preferring the more helpful response selects one option, then the user preferring the more honest response must select its alternative. In reality, the underlying factors driving human decisions may exhibit strong correlations. For instance, if an answer is helpful, it is also likely that it is honest. As a result, real-world datasets do not consist of just conflicting examples despite being generated by distinct reward functions. In this section, we investigate how correlations within the collected datasets of preference pairs affect the model's ability to identify a user's underlying reward function correctly. We begin with an analytics example inspired by the helpful-honest dataset.

\begin{example}\label{example:correlation}
    Suppose all options $y \in \gY$ can be represented with two numbers $y = [h_0, h_1]$ (e.g. helpful and honest scores). We assume that our population of users consists of two equally sized groups of users represented by ground-truth latent variables $z^* \sim \textrm{Bernoulli}(0.5)$; with the corresponding reward functions $r(y, z^*=0) = h_0$ and $r(y, z^*=1)= h_1$. Decisions are taken according to:
    $$p(y^w \succ y^\ell \vert z^*) = \mathbbm{1} \left\{r(y^w, z^*)   > r(y^\ell, z^*)\right\}$$
    This setup simulates a scenario where half of the users always prefer the more helpful while the half always choose the more honest answer. To investigate the impact of correlations in honest vs. helpful scores on the performance of our model, we will assume that scores of all options are i.i.d and normally distributed so that $[h_1, h_2] \sim \gN(\mu, \Sigma)$.
\end{example}

Suppose we observe a contextual dataset $\gD^C$ provided by a single user according to the abovementioned process. We wish to investigate the performance of a Bayes optimal classifier for determining the underlying value of $z^*$. We define 
$$\hat{Z}(\gD^C) = \argmax_{k \in \{0, 1\}}\sP(z^* = k \vert \gD^C).$$

We look at the error rate of $\hat{Z}$ defined as:

\begin{equation}
    \gE := \E\left[\hat{Z}(\gD^C) \neq z^*\right].
\end{equation}

In the above, the expectation is taken over all possible observable contexts $\gD^C$. In case when we restrict $\gD^C$ to observations consisting of only a single preference choice $\gD^C = \{y^w \succ y^\ell\}$, and assuming that $\Sigma = \left[\begin{matrix}
    1 & \rho \\
    \rho & 1
\end{matrix}\right]$ 
we can show that $\gE = \frac{1}{4} + \frac{1}{2\pi}\arcsin(\rho)$ (see Appendix~\ref{app:proof}). Note, at $\rho = -1.0$, the dataset consists of 100\% conflicting pairs, resulting in the theoretical 0\% error rate, meaning that a single preference choice is sufficient to determine the underlying value of $z^*$ unambiguously. On the other hand, at $\rho = 1.0$, none of the preference pairs are conflicting. If a response is helpful, it must also be honest. Such data does not allow us to differentiate between the underlying reward functions of individuals, resulting in an expected error rate of 50\%.

\begin{figure}[h]
    \centering
    \includegraphics[width=0.95\linewidth]{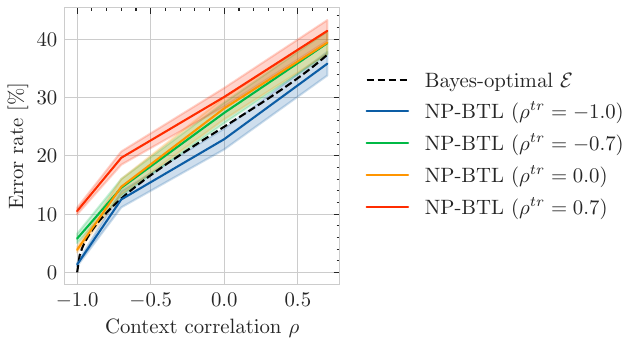}
    \caption{\textit{Error rate of NP-BTL vs Bayes optimal.}}
    \label{fig:bayes-optimal}
\end{figure}

In light of the above, we now assess the performance of our NP-BTL model on synthetic data generated according to the process described in Example~\ref{example:correlation}. We fit four independent NP-BTL models where, at training time, the correlation $\rho$ is set to $\rho^{tr} \in \{-1.0, -0.7, 0.0, 0.7\}$. This corresponds to a rate of conflicting pairs of 100\%, 75\%, 50\% and 25\%, respectively. Figure \ref{fig:bayes-optimal} shows the error rate of all four NP-BTL models where the two options forming the single preference pair provided in context are sampled from a normal distribution with a varying parameter $\rho$ value. We find that all models achieve near-optimal error rates, with the error rate increasing as the correlation in the training set increases. This confirms our hypothesis that datasets exhibiting strong correlations provide weaker training signals for the NP-BTL model to learn how to differentiate between conflicting reward functions of individuals. 

\subsubsection{Helpfulness vs. Honesty}

\begin{figure}[h]
    \centering
    \includegraphics[width=\linewidth]{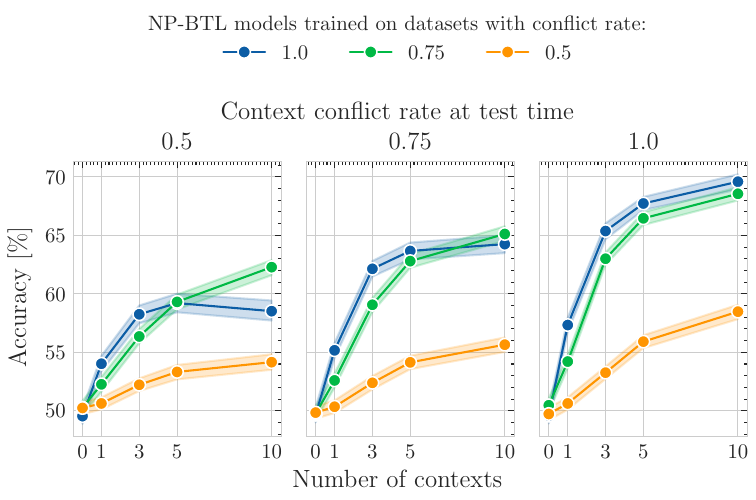}
    \caption{\textit{Accuracy of NP-BTL rewards vs. conflict rate on the HH dataset.}}
    \label{fig:hh-conflict-rate}
\end{figure}

We now go back to the Helpfulness vs. Honesty dataset introduced in section~\ref{sec:exp-hh} to investigate the impact of the conflict rate within the training dataset and in-context examples on the performance of our NP-BTL rewards. We train three NP-BTL models on three versions of the HH dataset, with rates of conflicting pairs of 0.5, 0.75 and 1.0, both in the context and target pairs. We evaluated these models on a held-out test set where the rate of conflicting pairs in $\gD^C$ varies, and the rate of conflict in $\gD^T$ is set to 1.0. \ref{fig:hh-conflict-rate} shows the results. We observe that model performance decreases for all models as the in-context conflict rate decreases. Contextual pairs which are not conflicting do not provide the necessary information to identify the hidden intentions of the user. We also observe that the performance of the NP-BTL model trained on data with a 0.5 conflict rate is significantly worse than that of the NP-BTL models trained on datasets with conflict rates of 1.0 and 0.75. This confirms our hypothesis from earlier - data with low conflict rates may not provide sufficiently strong training signals to learn to identify distinct human preferences.

\section{Related Work}

\textbf{RLHF and DPO.} Reinforcement Learning from Human Feedback (RLHF) has been foundational to aligning AI systems with human preferences. This technique typically involves learning a reward function from binary human preference data, commonly modelled using the BTL framework \cite{bradley_rank_1952}. The applications of RLHF to language models has been a significant focus of recent research, aiming to align LLMs with human values and preferences. Notable works in this area include those focusing on fine-tuning models \cite{bai_constitutional_2022, stiennon_learning_nodate, kim_preference_2022, ouyang_training_2022}. However, aligning LLMs through RLHF remains challenging due to training instability, which typically requires learning a reward model and the subsequent fine-tuning of the LLM. To address these inefficiencies, \cite{rafailov_direct_2024} propose Direct Preference Optimisation (DPO), a method that bypasses the need for explicit reward model training by optimising preference data directly in a supervised fashion.

\textbf{Pluralistic Alignment.} Most RLHF and DPO methods are based on the BTL model, which assumes a single reward function for all users. This approach can fail to capture the diversity of user preferences. \cite{sorensen_roadmap_2024} recognise, however, the need to focus on developing methods that enable pluralistic alignment of LLMs, serving humans with varying values and perspectives. 
A critical development in pluralistic alignment has been in recognising the so-called ``hidden context effect''--factors that influence user preferences in ways that are not directly observable. To address these effects, \cite{siththaranjan_distributional_2024} propose Distributional Preference Learning (DPL), modelling a distribution over the plausible reward values for each competing option. While DPL offers a more nuanced understanding of how hidden contexts influence the population's preferences, it does not provide a way of adapting the reward functions, and thus the final LLM policies, to individual users. Our approach builds on the concept of steerable pluralism \cite{sorensen_roadmap_2024}, wherein a model is steered such that its responses faithfully represent preferences of a given user or a subgroup of users. 

\textbf{Multi-objective Approaches.} Multi-objective learning frameworks are studied in the context of aligning AI systems with conflicting or diverse user preferences. One popular approach is Pareto-optimal optimisation, which aims to balance different objectives by finding optimal solutions with respect to a finite set of criteria \cite{boldi_pareto-optimal_2024, zhong_panacea_2024}. Rather than aiming for a single Pareto-optimal policy, we focus on enabling a flexible, steerable adaptation of LLM policies to individual preferences. Some works have considered extensions of multi-objective fine-tuning to allow such forms of adaptation. For instance, Reward Soup~\cite{rame_rewarded_2023} is an RLHF method which trains multiple LMs--one per a single reward function--and linearly interpolates the parameters of the fine-tuned LLMs to enable conditional generation at inference time. Similarly, \cite{wang_conditioned_2024}, given a fixed set of reward functions, proposes to train a fixed set of adapting parameters, which, combined with the base-model parameters, enable conditional generation. Another line of work \citep{wang_arithmetic_2024, guo_controllable_2024} considers prompt-based approaches that finetune an LM that is steered by simply providing the reward weightings as text within the prompt.
In contrast to these approaches, our method does not require access to datasets labelled with respect to a fixed set of pre-defined objectives, nor does it require the user to express their preferences as a vector of weights on a set of pre-defined objectives. Instead, our training data consists of a heterogeneous mixture of preferences expressed by multiple users having potentially conflicting objectives. Instead of describing the internal preferences of a user through a fixed vector of weights, our framework infers the user-specific reward function based on a small, few-shot sample of their preference choices. Our approach can be seen as a generalisation of the concurrent work of \cite{poddar_personalizing_2024}, which models reward functions with a variational encoder conditioned on a fixed number of contextual preference pairs. We propose a method for conditioning the rewards on an arbitrary number of contextual data points. We suggest an extension to DPO that enables the training of LLM policies that can be flexibly adapted to individual preferences at inference time.

\section{Limitations}

Due to computational limitations, our experimental evaluation is limited to the simplistic UltraFeedback dataset and a relatively small LLM. To ensure safe and reliable deployment of few-shot steerable LLM policies with NP-DPO, experiments on larger datasets and 
LLMs with more parameters are required. We also note that the proposed modulation operation based on FiLM layers is just a prototype solution which may be further improved by considering more parameter-efficient approaches. 

\section{Conclusions}

This work introduces a novel framework for few-shot steerable alignment, addressing the challenge of personalising AI systems to diverse user preferences. By extending the BTL model to handle heterogeneous preferences with unobserved variability factors, we enable the inference of user preferences from small samples of their choices. Our conditional reward modelling approach (NP-BTL) and its extension to DPO (NP-DPO) allow for training steerable LLM policies that can adapt to individual user preferences with minimal data. Experiments on synthetic and real-world datasets demonstrate the effectiveness of our methods in efficiently capturing and aligning with diverse human preferences. We also analyse the value of contextual data in determining the underlying reward function of a user. Noting that not all preference pairs are equally informative, future research directions include exploring active learning approaches for heterogeneous preference data collection, enabling maximum information gain about the users' internal preferences. Another line of future work should consider scaling the framework to larger models and more complex datasets. In summary, this paper's proposed methods and insights mark a significant step towards developing truly personalisable AI agents.

\section{Acknowledgements}
This work was supported by Azure sponsorship credits granted by Microsoft’s AI for Good Research Lab. Katarzyna Kobalczyk is supported by funding from Eedi, Claudio Fanconi is sponsored by Canon Medical Research and Hao Sun is sponsored by the Office of Naval Research.

\bibliography{references2, references3}
\bibliographystyle{icml2025}

\newpage
\appendix
\onecolumn

{\Large\bf Appendix}
\vspace{2mm}

\textbf{Code:} Code implementing the NP-BTL and NP-DPO models, alongside the instructions to reproduce the key experiments presented in this paper are made available at: \href{https://github.com/kasia-kobalczyk/few-shot-steerable-alignment}{https://github.com/kasia-kobalczyk/few-shot-steerable-alignment}.

\section{Analytic study: proof}\label{app:proof}

Suppose $y=[h_0, h_1] \sim \gN(\mu, \Sigma)$, where $\Sigma=\left[\begin{matrix}1 & \rho \\ \rho & 1\end{matrix}\right]$.  Let $Z^* \sim \mathrm{Bernoulli}(0.5)$, and
\begin{equation}
    r(y; Z^*) = \begin{cases}
        h_0 & \text{ if  } Z^* = 0 \\
        h_1 & \text{ if  } Z^* = 1
    \end{cases}
\end{equation}
We assume that decisions are made according to 
\begin{equation}
\sP(y_1 \succ y_2) = \begin{cases}
    1 & \text{ if } r(y_1; Z^*) > r(y_2; Z^*) \\
    0  & \text{ if } r(y_1; Z^*) < r(y_2; Z^*)
\end{cases}    
\end{equation}

We define a Bayes optimal classifier predicting the value of $Z^*$, given an observed dataset $\gD^C$ as :
\begin{equation}
    \hat{Z}(\gD^C) = \arg\max_{k \in \{0, 1\}}\sP(Z^*=k \vert \gD^C),
\end{equation}
where in case $\sP(Z^*=0 \vert \gD^C) = \sP(Z^*=1 \vert \gD^C)$, we let $\hat{Z}(\gD^C) \sim \mathrm{Bernouli}(0.5)$.

We look at the expected error rate of $\hat{Z}(\gD^C)$. We have that:

\begin{align*}
    \E\left[\hat{Z}(\gD^C) \neq Z^* \right] &= \sP\left(\hat{Z}(\gD^C) \neq Z^*\right) \\
    &= \sP\left(\hat{Z}(\gD^C) = 1 ,  Z^* = 0 \right) + \sP\left(\hat{Z}(\gD^C) = 0 ,  Z^* = 1 \right) \\ 
    &= 2\sP\left(\hat{Z}(\gD^C) = 1 ,  Z^* = 0 \right)  & (\text{by symmetry})\\
    &= \sP\left(\hat{Z}(\gD^C) = 1  \vert Z^* = 0 \right)
\end{align*}

Let us now consider a special case of a single contextual sample, such that $\gD^C = \{(y^w, y^\ell)\}$, where $y^w= [h_0^w, h_1^w]$, $y^\ell = [h_0^\ell, h_1^\ell]$ are sampled i.i.d from $\gN(\mu, \Sigma)$. We then have, for $k \in \{0, 1\}$

\begin{align}
    \sP\left(Z^*=k \vert \gD^C\right) \propto \sP(y^w \succ y^\ell \vert Z^* = k)\sP(Z^*=k) 
\end{align}
Therefore,
\begin{align*}
    \sP\left(Z^*=0 \vert \gD^C\right) = \begin{cases}
        0.5 & \text{ if } h_0^w > h_0^\ell \text{ and } h_1^w > h_1^\ell\\
        1.0 & \text{ if } h_0^w > h_0^\ell \text{ and } h_1^w < h_1^\ell \\ 
        0  & \text{otherwise}
    \end{cases}
    \quad 
    \sP\left(Z^*=1 \vert \gD^C\right) = \begin{cases}
        0.5 & \text{ if } h_1^w > h_1^\ell \text{ and } h_0^w > h_0^\ell\\
        1.0 & \text{ if } h_1^w > h_1^\ell \text{ and } h_0^w < h_0^\ell \\ 
        0  & \text{otherwise}
    \end{cases}
\end{align*}

The event $\hat{Z}(\gD^C) = 1$ can only happen in two cases: a) it happens with a probability of 1 if $\sP(Z^*=1 \vert \gD^C) > \sP(Z^*=0 \vert \gD^C)$ b) it happens with a probability of 0.5 when $\sP(Z^*=1 \vert \gD^C) = \sP(Z^*=0 \vert \gD^C)$. Conditioned on $Z^*=0$, we must necessarily have $h_0^w > h_0^\ell$. This in turn implies that $\hat{Z}(\gD^C) = 1$ can only happen in case b), i.e. when $\sP(Z^*=1 \vert \gD^C) = \sP(Z^*=0 \vert \gD^C) = 0.5$, which happens if and only if $h_1^w > h_1^\ell$ and $h_0^w > h_0^\ell$. Thus,
\begin{equation}
    \sP\left(\hat{Z}(\gD^C) \neq Z^*\right) = \frac{1}{2}\sP(h_1^w > h_1^\ell \vert h_0^w > h_0^\ell)
\end{equation}

To compute this probability we define $X = [X_0, X_1] = [h_0^w - h_0^\ell, h_1^w - h_1^\ell]$. We have that $X \sim \gN(\boldsymbol{0}, \Sigma)$ and our probability of interest is equal to 

\begin{equation}
     \sP\left(\hat{Z}(\gD^C) \neq Z^*\right) = \frac{1}{2}\sP(X_1 > 0 \vert X_0 > 0) = \sP(X_1 > 0, X_0 > 0),
\end{equation}

which by a standard result based on the geometry of the Gaussian distribution is equal $\frac{1}{4} + \frac{1}{2\pi}\arcsin(\rho)$.

\hfill\qedsymbol{}

\section{Complexity Analysis of NP-BTL and NP-DPO}\label{app:algorithm_analysis}

\textbf{Training.} Several operations influence the time complexity per iteration of the algorithm. First, sampling a user \(i\) from the dataset is an \(O(1)\) operation. Sampling the context set \(\gD_i^C\) and the target set \(\gD_i^T\) from the user-specific dataset \(\gD_i\) requires again \(O(1)\). During training, the number of context data points $N^C_i$ for each task varies between $N^{C, min}$ and  $N^{C, max}$. The number of target data points $N^T$ is constant for all tasks. 

At each forward pass, the algorithm computes the conditional latent variable $z^C_i$ for the sampled context datasets $\gD^C_i$. This operation takes $O(N^C_i \cdot f_{enc})$ where $f_{enc}$ is the complexity of a forward pass of the encoder network. The loss, i.e. negative log-likelihood of the target pairs $\gD^T_i$, is computed with a forward pass of all target pairs $(y^w_{i,j}, y^\ell_{i,j}) \in \gD_i^T$ and the latent variable $z_i^C$ through the decoder network. This takes $O(N^T \cdot f_{dec})$, where $f_{dec}$ is the complexity of one forward pass through the decoder network. 

Parameters of the model are optimised with a version of stochastic gradient descent run for a maximum of $T$ steps. Thus, the overall time complexity of training an NP-BTL or NP-DPO model is:
\[
O(T \cdot (N^{C,max}\ \cdot f_{enc} + N^T \cdot f_{dec} )),
\]
which is linear in the number of iterations \(T\), the number of target pairs \(N^T\), and the maximum number of context pairs \(N^{C,max}\).

\textbf{Inference.} To infer the reward for a single option $y \in \gY$, given a contextual sample $\gD^C_i = \{(y^w_{i,j}, y^\ell_{i,j})\}_{j=1}^{N^C_i}$, the model first computes the contextual embedding $z^C_i$ and then passes it through the decoder network together with the option $y$. Thus, time complexity of inference is $O(N^C_i \cdot f_{enc} + f_{dec})$.

\section{Implementation and experimental details}
All experiments are implemented in Python, using the PyTorch~\cite{paszke2017automatic} and HuggingFace libraries \cite{wolf2020huggingfacestransformersstateoftheartnatural}. All experiments can be run on a machine with a single A100 Nvidia GPU. We provide the code implementing the NP-BTL and NP-DPO models alongside the instructions needed to reproduce the key experiments in this repository: \href{https://github.com/kasia-kobalczyk/few-shot-steerable-alignment}{https://github.com/kasia-kobalczyk/few-shot-steerable-alignment}.

All reward and LLM policies are trained using the Adam optimiser~\cite{adam}. 

Unless otherwise stated, all mentions of MLPs are implemented as a two-layer neural networks with a hidden dimension of 256 and GELU activation.

Following the convention of NP training \cite{garnelo_conditional_2018, garnelo_neural_2018}, all tasks $\tau = (\gD^C, \gD^T)$ are constructed such that context pairs are a subset of the target pairs, i.e. $\gD^C \subseteq \gD^T$. We only report the unseen pairs' accuracies during model evaluation: $\gD^T \setminus \gD^C$.

\subsection{Illustrative comparison of existing approaches}\label{sec:appdx-illust-details}

\textbf{Dataset.}  In this experiment all options $y \in \gY$ are represented in $\gY = [-1, 1]$. The dataset is constructed by first generating 20k pairs sampled uniformly from $[-1, 1]^2$. This dataset is split into 10k training, 5k validation and 5k testing pairs. To construct a single task $\tau_i = (\gD^C_i, \gD^T_i)$, we first sample the unobservable $z^*_i \sim \mathrm{Bernoulli}(0.5)$. Then, we sample $N^C_i$ context and $N^T$ target pairs from the training, validation, or testing split. For each pair in the context and target datasets, the preference choices are determined deterministically so that $p(y_1 \succ y_2) = \mathbbm{1}\{r(y_1, z^*) > r(y_2, z^*)\}$, where
\begin{align}
    r(y, z^*) &= x  \ \text{ for } \ y \leq 0.5 
    \quad  \text{and} \\ 
    r(y, z^*) &= \begin{cases}
        2y & \text{if } z^* = 0 \\
        1 - y & \text{if } z^* = 1
    \end{cases} 
    \ \text{ for } y > 0.5
\end{align}
The number of context pairs varies uniformly between $N^{C,min} = 0$ and $N^{C,max} = 10$. The number of target pairs is set to $N^T = 20$. 

\textbf{NP-BTL implementation.} The encoder network $h_{\phi_e}$ is implemented as a DeepSet~\cite{zaheer_deep_2018}, with the inner and outer encoders implemented as MLPs: $h_{\phi_{e,1}} : \mathbb{R}^2 \rightarrow \mathbb{R}^{256}$, $h_{\phi_{e,2}} : \mathbb{R}^{256} \rightarrow \mathbb{R}^{256}$. The inputs to the inner network are preference pairs from the context dataset $(y^w, y^\ell) \in \gD^C$ concatenated together. The decoder $g_{\phi_{d}} : \mathbb{R}^{256 + 1} \rightarrow \mathbb{R}$, first maps a single target option $y \in \gD^T$ to a 256-dimensional embedding $e_y$ with an MLP. Then, $e_y$ is concatenated with the embedding of the context dataset, $z^C = h_\phi(\gD^C)$, and passed through another MLP to obtain the value of the conditional reward $r_\phi(y \vert \gD^C_i) \in \mathbb{R}$, i.e.:
$$r_\phi(y \vert \gD^C_i) = g_{\phi_d}(y, z^C_i) = \mathrm{MLP}_2\left(\left[ \mathrm{MLP}_1(y) || z_i^C \right]\right)$$

\textbf{Baseline implementation.} Both the BTL and DPL models are implemented as MLPs. For the BTL model, the 1-dimensional output represents the reward value of a single option $r_\phi(y)$. For the DPL model, the output is 2-dimensional, representing the mean and the logarithm of the standard deviation of a normal distribution $\gN(\mu(y), \sigma(y))$, describing the distribution of the likely values that $r_\phi(y)$ takes.

\textbf{Training.} Each batch consists of 64 tasks containing the same number of context data points  ($\gD^C_i, \gD^T_i)$ (for the non-conditional baselines we have $\gD^C_i = \varnothing$). Models are trained for a maximum of 50 SGD updates, retaining the model parameters with the lowest loss on the validation split. The learning rate is set to 1e-4.

\subsection{Real-world example: Helpfulness vs. Honesty}\label{sec:appdx-hh-details}

\textbf{Dataset.} The basis of this experiment is the UltraFeedback\footnote{\href{https://huggingface.co/datasets/openbmb/UltraFeedback}{https://huggingface.co/datasets/openbmb/UltraFeedback}} dataset. Each option $y = [x, \tilde{y}]$ within this dataset is represented with a user prompt $x$ and a candidate LLM response $\tilde{y}$. Alongside the set of prompts and candidate responses, each response in this dataset is assigned a score in four categories: honesty, helpfulness, truthfulness and instruction following. To simulate users with varying preferences, we define two reward functions:
\begin{align}
    r(y,z^*) = 
    \begin{cases}
         \mathrm{honesty}(\tilde{y})  & \text{ if } z^* = 0 \\
         \mathrm{helpfulness}(\tilde{y}) & \text{ if } z^* = 1
    \end{cases}
\end{align}
The raw format of prompts and responses is in natural language. For computational efficiency, we represent them as frozen embeddings pre-computed with the \texttt{LLama-3-8B}\footnote{\href{https://huggingface.co/meta-llama/Meta-Llama-3-8B}{https://huggingface.co/meta-llama/Meta-Llama-3-8B}} language model. These embeddings are obtained as the last hidden state representation of the last token in the tokenized representation of $y$. We denote the embeddings of $y_{i,j}^w, y_{i,j}^\ell$ as $e_{i,j}^w, e_{i,j}^\ell \in \mathbb{R}^{4096}$. During training, validation, and testing, we first sample $z^*_i \sim \mathrm{Bernoulli}(0.5)$ and then sample $N^C_i$ context pairs and $N^T$ target pairs with decisions made deterministically according to $p(y_1 \succ y_2) = \mathbbm{1}\{r(y_1, z^*_i) > r(y_2, z^*_i)\}$. 

During training, the number of context pairs for reward modelling varies uniformly between $N^{C,min} = 0$ and $N^{C,max} = 10$. The number of target pairs is set to $N^T = 20$.  

For LLM policy fine-tuning, during training, the number of context pairs varies uniformly between $N^{C,min} = 0$ and $N^{C,max} = 6$. The number of target pairs is set to $N^T = 9$. 
 
To construct the dataset with 100\% conflicting pairs, we filter the dataset to pairs of options for which the response made with $r(\cdot; 0)$ differs from the response made with $r(\cdot; 1)$. As a result, our final dataset consists of 28763 training, 3673 validation, and 3509 testing pairs $(y_1, y_2)$. 

\subsubsection{Reward Models}

\textbf{NP-BTL.} Our context encoder is implemented as a DeepSet with multi-head self-attention. For $\gD^C_i = \{(y_{i,j}^w, y_{i,j}^\ell)\}_{j=1}^{N^C_i}$ the contextual embedding $z^C_i$ is computed as:
\begin{equation}
     z_i^C = h_{\phi_{e, 2}}\left(\sum_{j=1}^{N^C_i} \text{MultiHeadAttn}(W^T[e_{i,j}^w || e_{i,j}^{\ell}])\right),
\end{equation}
where $W \in \R^{2\cdot4096  \times 256}$ is a linear map, $h_{\phi_{e, 2}}: \mathbb{R}^{256} \rightarrow \mathbb{R}^{256}$ is an MLP,  and $\text{MultiHeadAttn}$ a multi-head attention layer with 8 heads and dropout 0.1 \footnote{\href{https://pytorch.org/docs/stable/generated/torch.nn.MultiheadAttention.html}{https://pytorch.org/docs/stable/generated/torch.nn.MultiheadAttention}}.

The decoder $g_{\phi_{d}} : \mathbb{R}^{256 + 256} \rightarrow \mathbb{R}$, first passess the embedding $e$ of a target option $y \in \gD^T$ through an MLP, then concatenates the output with $z_i^C$, and passess it through another MLP so that:
\begin{equation}
r_{\phi}(y | \gD_i^C) =  g_{\phi_d}(y, z^C_i) = \mathrm{MLP}_2\left(\left[ \mathrm{MLP}_1(e)|| z^C_i\right]\right)
\end{equation}

\textbf{BTL.} The BTL model is a simple MLP mapping the embedding $e \in \mathbb{R}^{4096}$ to scalar rewards in $\mathbb{R}$.

\textbf{Training.} Each batch consists of 64 tasks containing the same number of context data points  ($\gD^C_i, \gD^T_i)$ (for the non-conditional baselines we have $\gD^C_i = \varnothing$). Models are trained for a maximum of 450 SGD updates, retaining the model parameters with the lowest loss on the validation split. The learning rate is set to 1e-4.

\subsubsection{LLM Policy Models}\label{app:dpo_implementation}

\textbf{NP-DPO.} Contextual embeddings $z^C_i$ are obtained in the same way as in reward modelling with NP-BTL. As the reference model for policy training, we choose the \texttt{gemma-2b} LLM\footnote{\href{https://huggingface.co/google/gemma-2b}{https://huggingface.co/google/gemma-2b}}, as it is small enough to be trained on a single A100 GPU. A key component in this task is modulating the LLM parameters $\theta$ with the conditional latent variable $z_i^C$. Our modulation operator is based on feature-wise linear modulation (FiLM)~\cite{perez_film_2017}. The modulating hypernetwork $h_\phi$, takes in the embedding $z_i^C$ and outputs a collection of modulating parameters $h_\phi(\gD^C_i) = \{\gamma_{m}(z_i^C), \beta_{m}(z_i^C)\}_{m=1}^L$, where $L$ is equal to the number of attention blocks. If we denote by $o_m$  the output of the attention block at the $m^{\text{th}}$ layer, then hidden representations modulated with $z_i^C$ are obtained as:
\begin{equation}
     o_m' = \gamma_m(z_i) \cdot o_m + \beta_m(z_i).
\end{equation}
Here, we implement $\gamma_m$ and $\beta_m$ as simple linear maps between $\sR^{256}$ and $\sR^{4096}$. The parameters $\theta$ of the base LLM and the collection of the weights $\phi$ defining the hyper network $h_\phi$ are optimised jointly by maximising the conditional DPO objective:
\begin{equation}
    \max_{\theta, \phi}\mathop{\mathbb{E}}_{\substack{(\gD^C_i, \gD^T_i),\\ (x, \tilde{y}^w, \tilde{y}^\ell) \in \gD^T_i}}\left[\log \sigma\left(\beta \log \frac{\pi_\theta(\tilde{y}^w \vert x, \gD^C_i)}{\pi_{\mathrm{ref}}(\tilde{y}^w \vert x)} -  \beta \log \frac{\pi_\theta(\tilde{y}^\ell \vert x, \gD^C_i)}{\pi_{\mathrm{ref}}(\tilde{y}^\ell \vert x)} \right)\right].
\end{equation}

\textbf{BTL-DPO.} We optimise the parameters $\theta$ of \texttt{gemma-2b} with the standard DPO objective:
\begin{equation}
        \max_{\theta} \mathop{\E}_{(x, \tilde{y}^w, \tilde{y}^\ell)}\left[\log \sigma\left(\beta \log \frac{\pi_\theta(\tilde{y}^w \vert x)}{\pi_{\mathrm{ref}}(\tilde{y}^w \vert x)} - \beta \log \frac{\pi_\theta(\tilde{y}^\ell \vert x)}{\pi_{\mathrm{ref}}(\tilde{y}^\ell \vert x)}\right)\right].
\end{equation}

\textbf{Training.} To reduce memory requirements, parameters $\theta$ of the base \texttt{gemma-2b} LLM are optimised with Low-Rank Approximation (LoRA)~\cite{hu_lora_2021}, with a rank of 512 and a LoRA-$\alpha$ of 1024. The learning rate of the Adam optimizer is set to 1e-6. The parameter $\beta$ is set to 0.05. During training, we use a batch size of 1.

\subsection{Correlations in data and identifiability of the hidden context}

\subsubsection{Analytic study}

\textbf{Dataset.} Options $y = [h_0, h_1] \in \sR^2$ are sampled with $\gN(\mu, \Sigma)$ with $\mu = \left[\begin{matrix} 0 \\ 0 \end{matrix}\right]$ and $\Sigma = \left[\begin{matrix} 1 & \rho \\ \rho & 1\end{matrix}\right]$. For any two pairs $y_1, y_2 \in \sR^2$, decisions are made according to $p(y_1 \succ y_2) = \mathbbm{1}\{r(y_1, z^*) > r(y_2, z^*)\}$, where
\begin{equation}
    r(y, z^*) = \begin{cases}
        h_0 & \text{ if } z^* = 0 \\
        h_1 & \text{ if } z^* = 1
    \end{cases}
\end{equation}
The dataset is constructed by first generating 20k pairs $y_1, y_2 \in \sR^2$. This dataset is split into 10k training, 5k validation and 5k testing pairs. To construct a single task $\tau_i = (\gD^C_i, \gD^T_i)$, we first sample the unobservable $z^*_i \sim \mathrm{Bernoulli}(0.5)$ and then sample $N^C_i$ context and $N^T$ target pairs from the training, validation, or testing split with preference choices made according to $r(\cdot \ ; z^*_i)$. The number of context pairs varies uniformly between $N^{C,min} = 0$ and $N^{C,max} = 10$. The number of target pairs is set to $N^T = 20$. 

\textbf{Model implementation and training.} Implementation of the NP-BTL and the baseline BTL reward models is analogous to that presented in \ref{sec:appdx-illust-details}, with the input dimension replaced from 1 to 2. The training setup is identical to \ref{sec:appdx-illust-details}. 

\subsubsection{Helpfulness vs. Honesty}

\textbf{Dataset.} The construction of the dataset is analogous to the one presented in \ref{sec:appdx-hh-details}, controlling the number of conflicting pairs. We construct three independent datasets with a conflict rate of 100\% (this is the same dataset as in \ref{sec:appdx-hh-details}), 75\% and 50\%. All datasets are divided into training, validation, and testing splits, with the same respective sizes as in the 100\% condlicting case. 

\textbf{Model implementation and training.} Identical to the setup described in \ref{sec:appdx-hh-details}.

\section{Additional Experiments}\label{app:additional_experiments}

\subsection{UltraFeedback with three Behavioural modes: Helpfulness vs Honesty vs Truthfulness}\label{app:HHT}

\textbf{Dataset.} We use the UltraFeedback dataset and follow an analogous setup to the one described in \ref{sec:appdx-hh-details}, this time sampling reward functions with three different modes of behaviour. That is we let:
To simulate users with varying preferences, we define two reward functions:
\begin{align}
    r(y,z^*) = 
    \begin{cases}
         \mathrm{honesty}(\tilde{y})  & \text{ if } z^* = 0 \\
         \mathrm{helpfulness}(\tilde{y}) & \text{ if } z^* = 1 \\
         \mathrm{truthfulness}(\tilde{y}) & \text{ if } z^* = 2
    \end{cases}
\end{align}
where $z^* \sim \mathrm{Multinomial}([1/3, 1/3, 1/3])$. To ensure high informativeness of contextual samples, we filter the dataset to pairs $y_1, y_2$ for which one choice is made according to $r(\cdot \ ; 0)$, $r(\cdot \ ; 1)$, $r(\cdot \ ; 2)$ is distinct from the others The resulting dataset consists of 27475 training, 3415 validation, and 3415 testing pairs.

\textbf{Training.} We train the BTL reward models and DPO policies with the same hyperparameters as in \ref{sec:appdx-hh-details}. As previously, we train two versions of the BTL models: on data generated according to the mixture of three reward functions (BTL-mixed) and one model for each of the three ground-truth values of $z^*$ (BTL-helpfulness, BTL-honesty, and BTL-truthfulness).

\begin{figure*}[h]
    \centering
    \subfloat[0.5\linewidth][
        Accuracies of the baseline BTL models (left) and the NP-BTL model (right).
        \label{fig:hht-reward}
    ]{
        \includegraphics[width=0.47\linewidth]{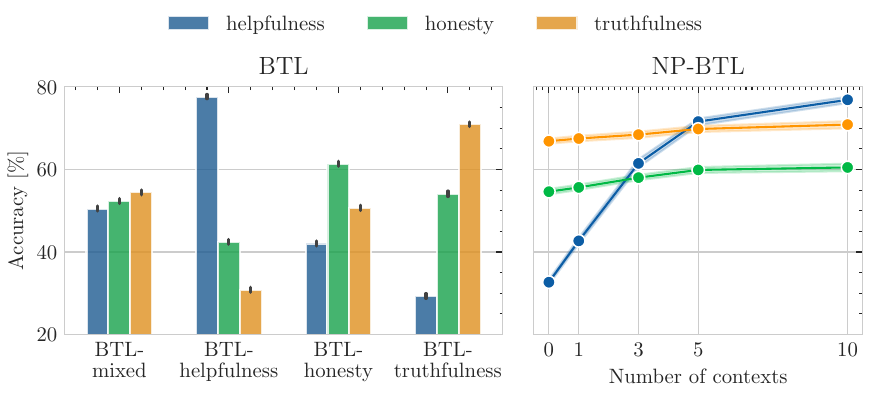}
    }
    \quad
    \subfloat[0.5\linewidth][
        Accuracies of the implicit rewards of LLM policies trained with BTL-DPO (left) and NP-DPO (right).
        \label{fig:hht-policy}
    ]{
        \includegraphics[width=0.47\linewidth]{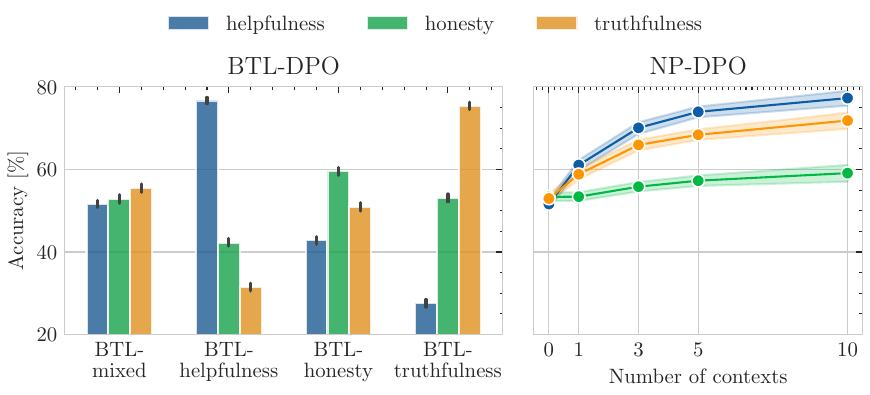} 
    }
    \caption{\textit{Performance of a) NP-BTL and b) NP-DPO vs. their non-conditional counterparts.}}
    \label{fig:hht-results}
\end{figure*}

\textbf{Results.} Figure~\ref{fig:hht-results} shows the test-set accuracies of the learned rewards evaluated separately on preference choices made with the helpfulness- ($r(\cdot \ ; 0)$ in blue), honesty- ($r(\cdot \ ; 1)$ in green), and truthfulness-preferring ($r(\cdot \ ; 2)$ in orange) reward functions. We make analogous conclusions to experiments on the 2-mode HH dataset from section 4.2 of the main paper. The BTL-mixed reward model fails to align with any of the three objectives. The accuracy of the NP-BTL model increases with the number of contextual preference pairs accross all three modes of behaviour. At $N^C_i = 10$, its performance matches the performance of the dedicated BTL-helpfulness, BTL-honesty and BTL-truthfulness models. 

\begin{figure}[h]
    \centering
    \vspace{-0.5em}
\includegraphics[width=0.95\linewidth]{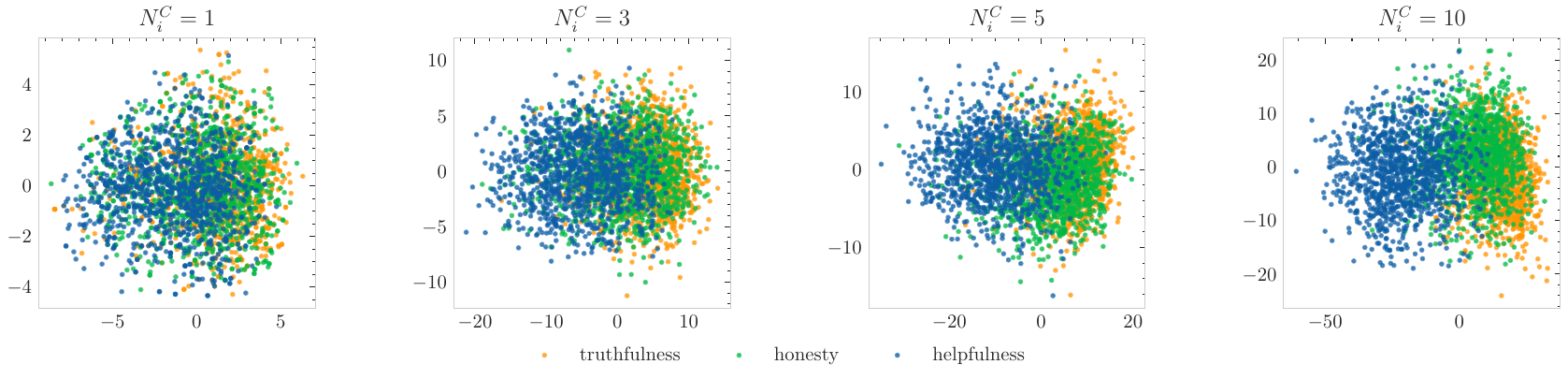}
    \caption{\textit{PCA of context embeddings on three behaviour modes}. Points are labelled by the true value of $z^*_i$.}
    \label{fig:hht-pca}\vspace{-0.5em}
\end{figure}

We also plot the PCA embeddings of the contextual latent variables $z_i^C$ (Figure~\ref{fig:hht-pca}). Again, with an increasing number of contextual data points, we observe $ z_i^C$ forming distinct clusters. The separation between honesty and truthfulness is less clearly defined. We hypothesise this is due to these two objectives being semantically similar, leading to choices made with $r(\cdot \ ; 1)$ (honesty) and $r(\cdot \ ; 2)$ (truthfulness) exhibiting a stronger correlation.

\subsection{Beyond a fixed number of reward functions}\label{app:continuous_IDs}

One of the appealing properties of the NP-BTL and NP-DPO models is that they are not constrained to modelling just a fixed set of reward functions. Instead, they can represent an entire \textit{spectrum} of behaviours. We illustrate this with the UltraFeedback dataset in a two-dimensional and a three-dimensional setup.

\textbf{2D setup.} We let $z^* = [z^*_0, z^*_1]^T \sim \mathrm{Dirichlet}([0.5, 0.5])$ and we let 
$$r_{2D}(y ; z^*) = z^*_0 \cdot \mathrm{helpfulness}(\tilde{y}) + z^*_1 \cdot \mathrm{honesty}(\tilde{y})$$

\textbf{3D setup.} We let $z^* = [z^*_0, z^*_1, z^*_2]^T \sim \mathrm{Dirichlet}([0.5, 0.5, 0.5])$ and we let 
$$r_{3D}(y ; z^*) = z^*_0 \cdot \mathrm{helpfulness}(\tilde{y}) + z^*_1 \cdot \mathrm{honesty}(\tilde{y}) + z^*_2 \cdot \mathrm{truthfulness}(\tilde{y})$$

We train one NP-BTL model for the 2D setup (NP-BTL-2D) and one for the 3D setup (NP-BTL-3D). Hyperparameters are set as in \ref{sec:appdx-hh-details}. 

\begin{figure*}[h]
    \centering
    \subfloat[0.32\linewidth][
        2-dimensional setup.
        \label{fig:hh-cont}
    ]{
        \includegraphics[width=0.32\linewidth]{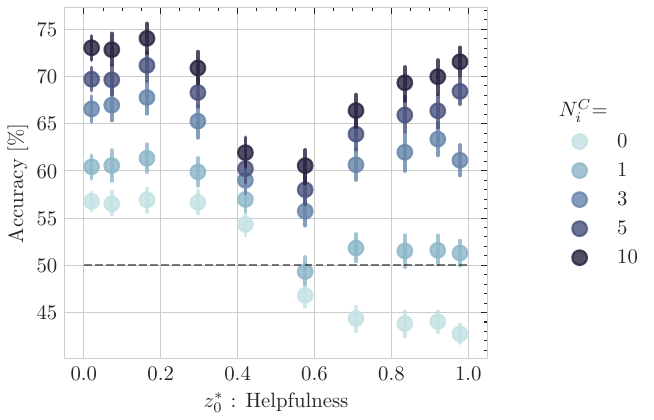}
    }
    \quad
    \subfloat[0.62\linewidth][
        3-dimensional setup.
        \label{fig:hht-cont}
    ]{
        \includegraphics[width=0.62\linewidth]{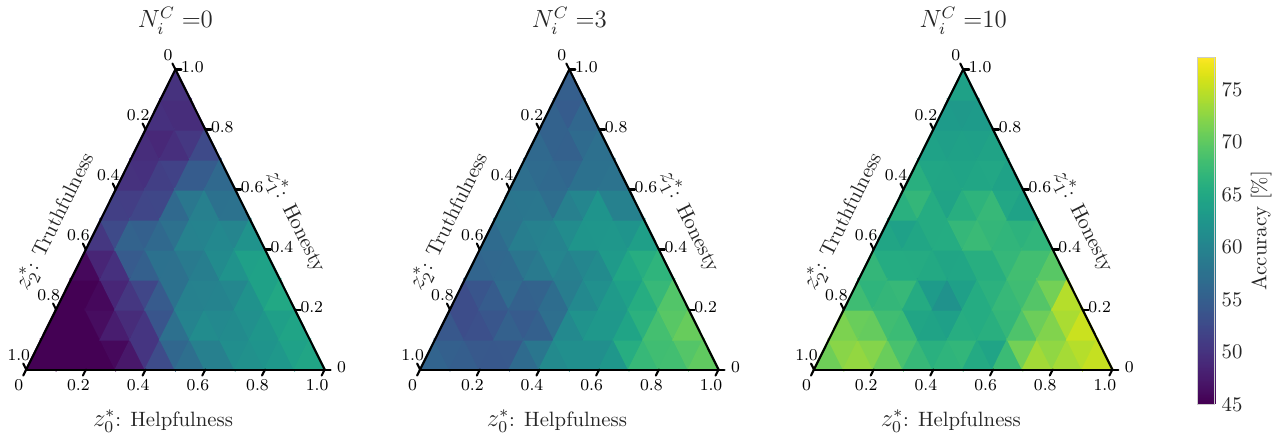}
    }
    \caption{\textit{Accuracies of the NP-BTL models.} a) $y$-axis represents the accuracies of the NP-BTL-2D rewards evaluated on test-set preference pairs with choices made according to $r_{2D}(\cdot \ ;z^*)$, where $z^*_0 = 1 - z^*_1$ varies between 0.0 and 1.0. Values of the $x$-axis are binned into ten equally-sized bins in the range $[0, 1]$. b) Colour corresponds to the accuracy of the NP-BTL-3D rewards evaluated on test-set preference pairs with choices made according to $r_{3D}(\cdot \ ; z^*)$, varying the value of $z^*$ inside the 3D simplex. The area of the simplex is divided into triangular bins with edge lengths of 0.1.}
    \label{fig:continuous-results}
\end{figure*}

\textbf{Results.} Figure~\ref{fig:continuous-results} shows the accuracies of the NP-BTL-2D and NP-BTL-3D rewards on the test-set pairs, with choices made according to $r_{2D}$ and $r_{3D}$, varying the value of the ground-truth $z^*$ inside the 2- and 3- dimensional simplex, respectively. Both in the 2D and 3D setups, with increasing contextual data points, the accuracy improves across the entire spectrum of behavioural modes represented by $z^*$.  This observation implies that our NP-BTL model successfully adapts its predictions to the observable contexts, enabling not only discrete, group-level alignment, but a truly personalisable, user-level alignment.

\end{document}